\documentclass{article}
\usepackage{ijcai22}

\usepackage{times}
\usepackage{soul}
\usepackage{url}
\usepackage[hidelinks]{hyperref}
\usepackage[utf8]{inputenc}
\usepackage[small]{caption}
\usepackage{graphicx}
\usepackage{amsmath}
\usepackage{amsthm}
\usepackage{booktabs}
\usepackage{multirow}
\usepackage{microtype}
\usepackage{wrapfig}
\usepackage{subcaption}
\usepackage{paralist} 
\usepackage[dvipsnames]{xcolor}
\usepackage{pifont}

\usepackage{fancyhdr}

\newcommand{\task}[1]{\textsf{\small #1}}

\usepackage[obeyFinal,textsize=scriptsize]{todonotes}

\title{
Knowing Earlier what Right Means to You: A Comprehensive VQA Dataset for Grounding Relative Directions via Multi-Task Learning}

\author{%
  Kyra Ahrens\footnote{Equal contribution.} \and Matthias Kerzel$^{*}$\and Jae Hee Lee$^{*}$\and Cornelius Weber \And Stefan Wermter%
  \affiliations%
  University of Hamburg%
  \emails%
  \{kyra.ahrens, matthias.kerzel, jae.hee.lee, cornelius.weber, stefan.wermter\}@uni-hamburg.de%
}

\begin{document}

\maketitle
\thispagestyle{fancy}

\begin{abstract}

Spatial reasoning poses a particular challenge for intelligent agents and is at the same time a prerequisite for their successful interaction and communication in the physical world. One such reasoning task is to describe the position of a target object with respect to the intrinsic orientation of some reference object via \textit{relative directions}. In this paper, we introduce  GRiD-A-3D, a novel diagnostic visual question-answering (VQA) dataset based on abstract objects. Our dataset allows for a fine-grained analysis of end-to-end VQA models' capabilities to ground relative directions. At the same time, model training requires considerably fewer computational resources compared with existing datasets, yet yields a comparable or even higher performance. 
Along with the new dataset, we provide a thorough evaluation based on two widely known end-to-end VQA architectures trained on GRiD-A-3D. We demonstrate that within a few epochs, the subtasks required to reason over relative directions, such as recognizing and locating objects in a scene and estimating their intrinsic orientations, are learned in the order in which relative directions are intuitively processed.

\end{abstract}

\section{Introduction}
\label{sec:introduction}

Reasoning to solve complex spatial tasks like grounding directional relations in an intrinsic frame of reference can be decomposed into a set of subtasks that are hierarchically organized. Consider two objects $o_1$ and $o_2$ in an image, where each of the objects has a clear front side and orientation. Learning to answer whether the triple $(o_1, r, o_2)$ holds for a given directional relation $r$ in a frame of reference that is intrinsic to $o_2$ spans the following stages (see Fig.\ \ref{fig:example} for an example):  

\begin{figure}[t!]
   \centering%
   \includegraphics[width=\linewidth]{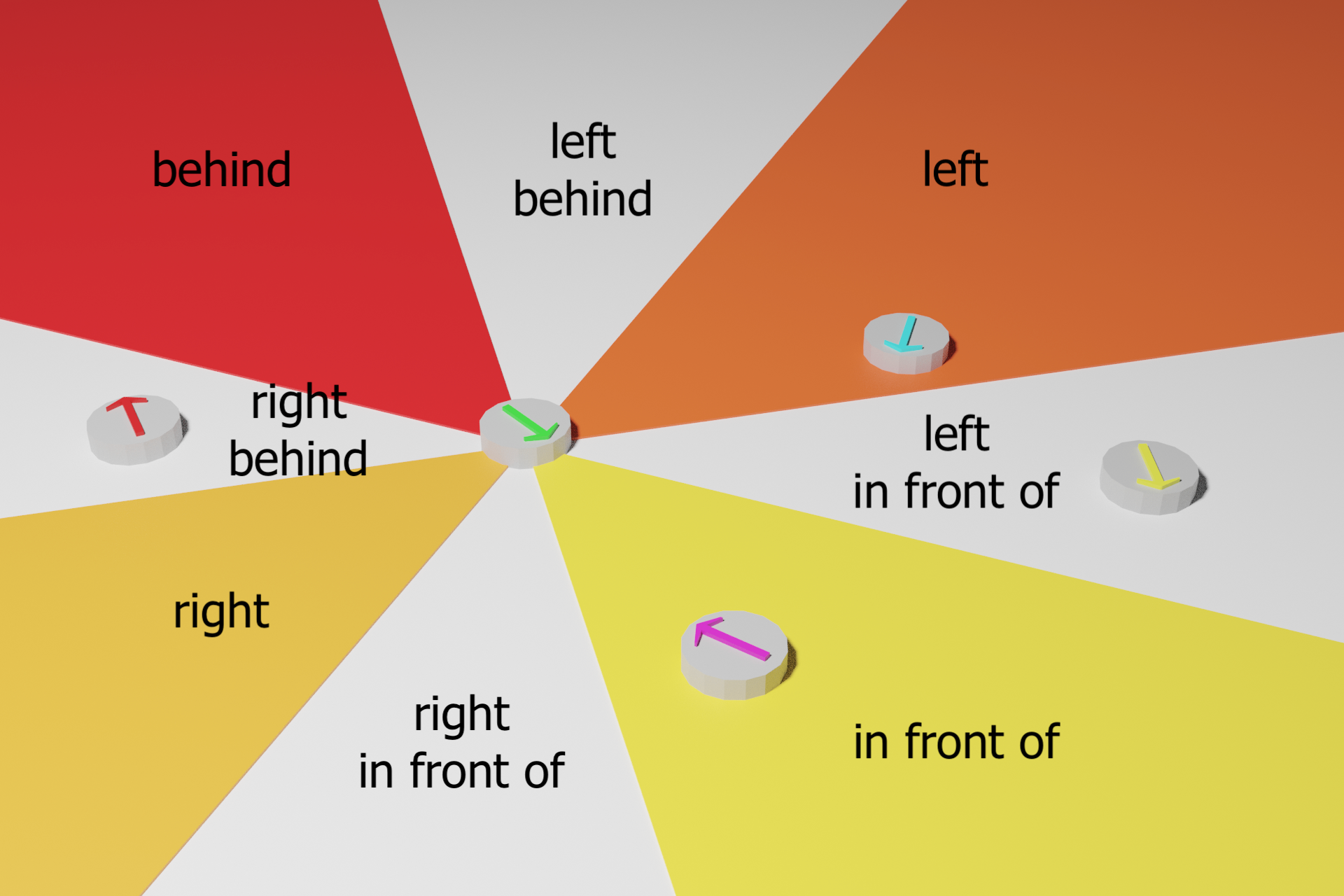}
   \caption{Example of grounding relative directions, e.g., considering the green arrow's perspective, the yellow arrow is on the left in front of it.}
   \label{fig:example}
 \end{figure}

\begin{enumerate}
\item Both the target object and the reference object have to be recognized in the image (\textbf{existence prediction}). In other words, an agent must initially be capable of answering questions such as ``Is $o_1$ in the image?'' or ``Is $o_2$ in the image?''. 

\item Next, the object's pose that defines the relative relation has to be discerned, enabling an agent to successfully respond to questions such as ``What is the cardinal direction of $o_2$?'' (\textbf{orientation prediction}).


\item Predicting the directional relation using the intrinsic frame of reference is learned by combining the two preceding competencies, allowing an agent to answer a question similar to ``What is the relation between $o_1$ and $o_2$ from the perspective of $o_2$?'' (\textbf{relation prediction}). Likewise, predicting which target object is in a specific relation to some reference object (\textbf{link prediction}) can be answered, e.g., ``Taking $o_2$'s perspective, which object is in relation $r$ to it?''.

\item Based on all previous stages, an agent can determine whether a specific directional relationship exists between the two objects (\textbf{triple classification}), thus successfully providing an answer to a question like ``From $o_2$'s perspective, is $o_1$ left of $o_2$?''.
\end{enumerate}

\begin{figure}[t!]
    \centering 
    \includegraphics[width=\linewidth]{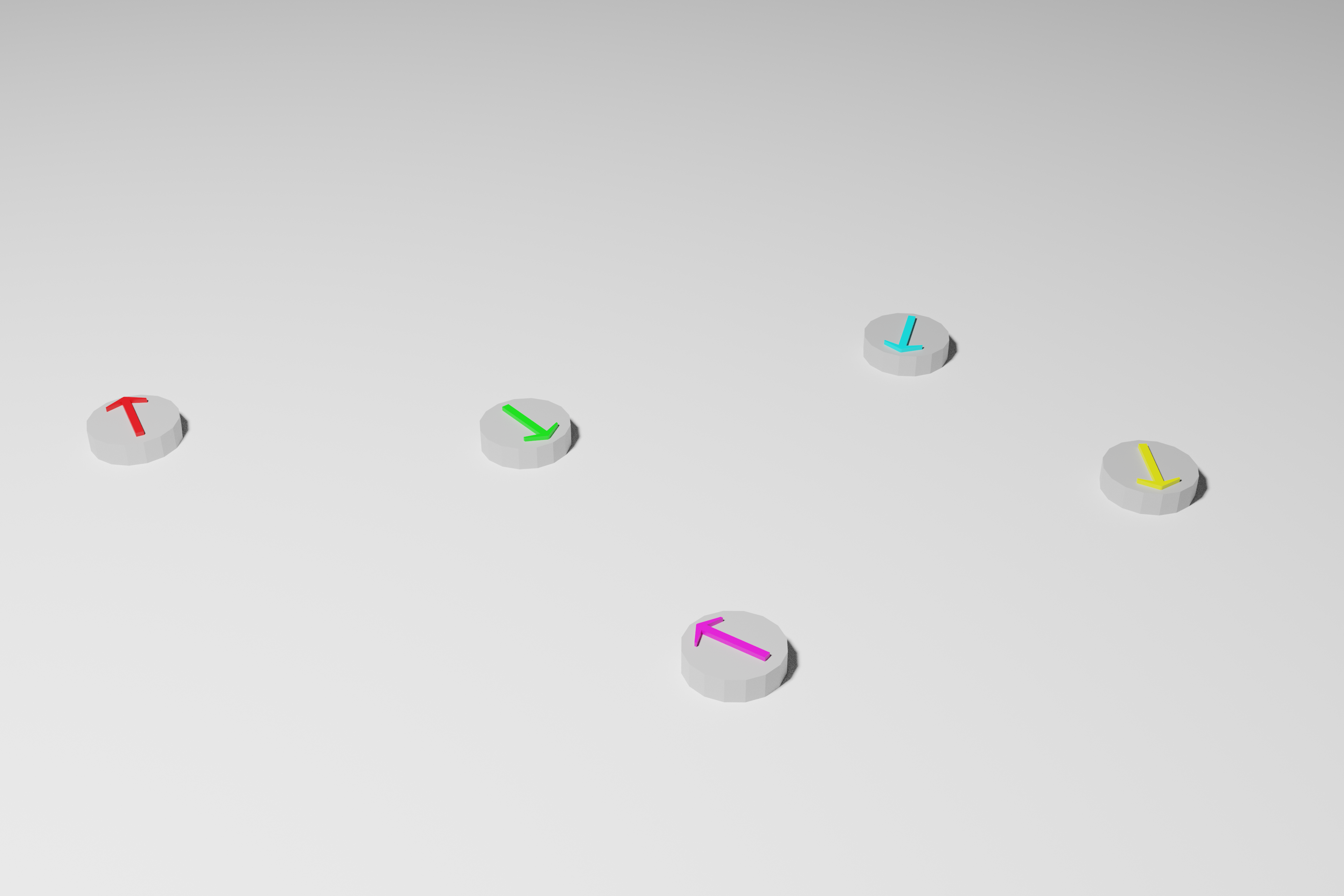}\\
    \vspace{0.2cm}
    \includegraphics[width=\linewidth]{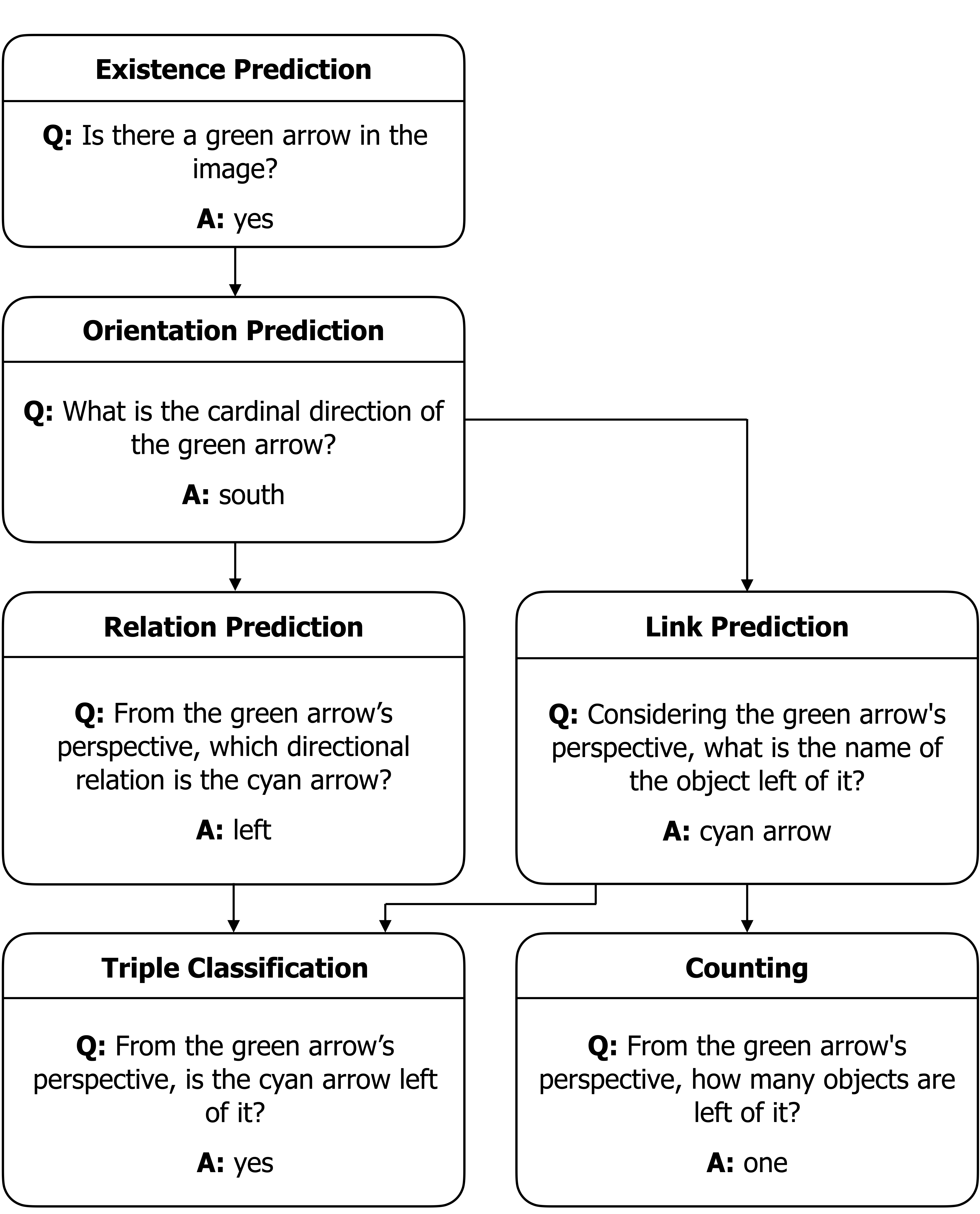}
    \caption{\textbf{Top}: Image from the GRiD-A-3D dataset. \textbf{Bottom}: Assumed hierarchy of spatial reasoning tasks to answer different question of the abstract GRiD-A-3D dataset. Arrows indicate a chronological dependency of tasks, e.g., in order to determine the orientation of an object, it first has to be recognized.}
   \label{fig:tasks}
 \end{figure}

\noindent In previous work~\cite{lee_grid3d_2022}, we showed that enabling a VQA architecture to reason about relative directions is viable, provided that all of the learning stages listed above are encapsulated in corresponding subtasks as summarized in Fig.\ \ref{fig:tasks}.
Beyond that, the following two observations were made: First, the subtasks that are found earlier in the chronology of learning stages are also learned earlier by the models, and second, this behavior is consistent for different neural end-to-end models. However, these findings are based on experiments involving images with 3D models of real objects, that may introduce a potential bias that confounds the analysis of reasoning about relative directions. 

In the present work, we introduce GRiD-A-3D, a novel and simplified diagnostic VQA dataset, which allows for a more efficient and targeted analysis of the corresponding reasoning process by removing possible biases from using real-world objects. Subsequently, we report the performance of the two established end-to-end VQA models MAC~\cite{hudson_compositional_2018} and FiLM~\cite{perez_film_2018} on this dataset. With our experiments, we show that, when trained on GRiD-A-3D, both models depict a similar qualitative learning behavior compared with their replica trained on the more complex non-abstract GRiD-3D~\cite{lee_grid3d_2022} dataset. At the same time, training converges up to three times faster, thus allowing more efficient neural experiments.


We summarize the contributions made in this paper as follows:
\begin{itemize}
    \item We complement our GRiD-3D benchmark suite\footnote{\url{https://github.com/knowledgetechnologyuhh/grid-3d}} with a novel GRiD-A-3D (\underline{\textbf{G}}rounding \underline{\textbf{R}}elat\underline{\textbf{i}}ve \underline{\textbf{D}}irections with \underline{\textbf{A}}bstract objects in \underline{\textbf{3D}}) dataset that enables a faster and less biased evaluation of spatial reasoning behavior in VQA compared with the original GRiD-3D dataset.
    \item We verify our previous research findings with the new dataset, thus underpinning our hypothesis that multi-task learning enables neural models to learn to ground relative directions in VQA.
    \item Furthermore, we add evidence to our hypothesis that during multi-task learning, spatial reasoning abilities of a neural model develop along the intuitive order of corresponding subtasks, thus forming an implicit curriculum.
\end{itemize}


\section{Related Work}
\label{sec:related-work}
Aiming to provide a suitable setup to assess the reasoning capabilities of neural models on vision-language tasks, diagnostic datasets have been introduced~\cite{johnson_clevr_2017,hudson_gqa_2019}. One of the major advantages of such datasets is that they provide structured and tightly controlled scenes to prevent models from circumventing reasoning by exploiting conditional biases that commonly arise with real-world images. A particular advantage of diagnostic datasets based on synthetic images is that their generation process is scalable, customizable, and therefore allows for a more fine-grained performance analysis. 

The vast majority of diagnostic VQA datasets is limited to spatial reasoning tasks based on the absolute frame of reference, i.e., object positions are relative to the viewer of the image. Yet taking into account more realistic scenarios such as multi-agent dialogue in a situated environment, understanding relative directions is a prerequisite for meaningful communication. As a consequence, early models to learn symbolic reasoning with relative directions have been proposed~\cite{moratz_spatial_2006,lee_starvars:_2013,hua_qualitative_2018}. However, they inherently assume the availability of scene annotations in terms of object labels and spatial relations instead of requiring a model to infer such information implicitly.

An early synthetic dataset providing a test bed for grounding relative directions is Rel3D~\cite{goyal_rel3d_2020}. Since Rel3D is restricted to two objects per scene and one single task, i.e., binary prediction of (object\textsubscript{1}, relation, object\textsubscript{2}) triples, GRiD-3D~\cite{lee_grid3d_2022} was introduced, which combines the advantage of a rich number of tasks and questions as found in traditional synthetic VQA datasets with the challenge of grounding relative directions. 

GRiD-3D is the first-of-its-kind to target multi-task learning of relative directions in a controlled setting. With this dataset, it was shown that, before learning how to answer the question whether a triple (object\textsubscript{1}, relation, object\textsubscript{2}) holds, neural end-to-end VQA models rely on an implicit curriculum of related subtasks such as object detection, orientation estimation, and relation prediction~\cite{lee_grid3d_2022}. Objects in GRiD-3D cover a variety of categories, ranging from humanoids and animals to furniture and vehicles. Naturally, such objects differ in terms of proportions, complexity, and, most importantly, symmetry, which can be a crucial determinant of how easily a neural network can infer their orientation (and perform associated tasks).


\begin{figure}[t!]
    \includegraphics[width=0.497\linewidth]{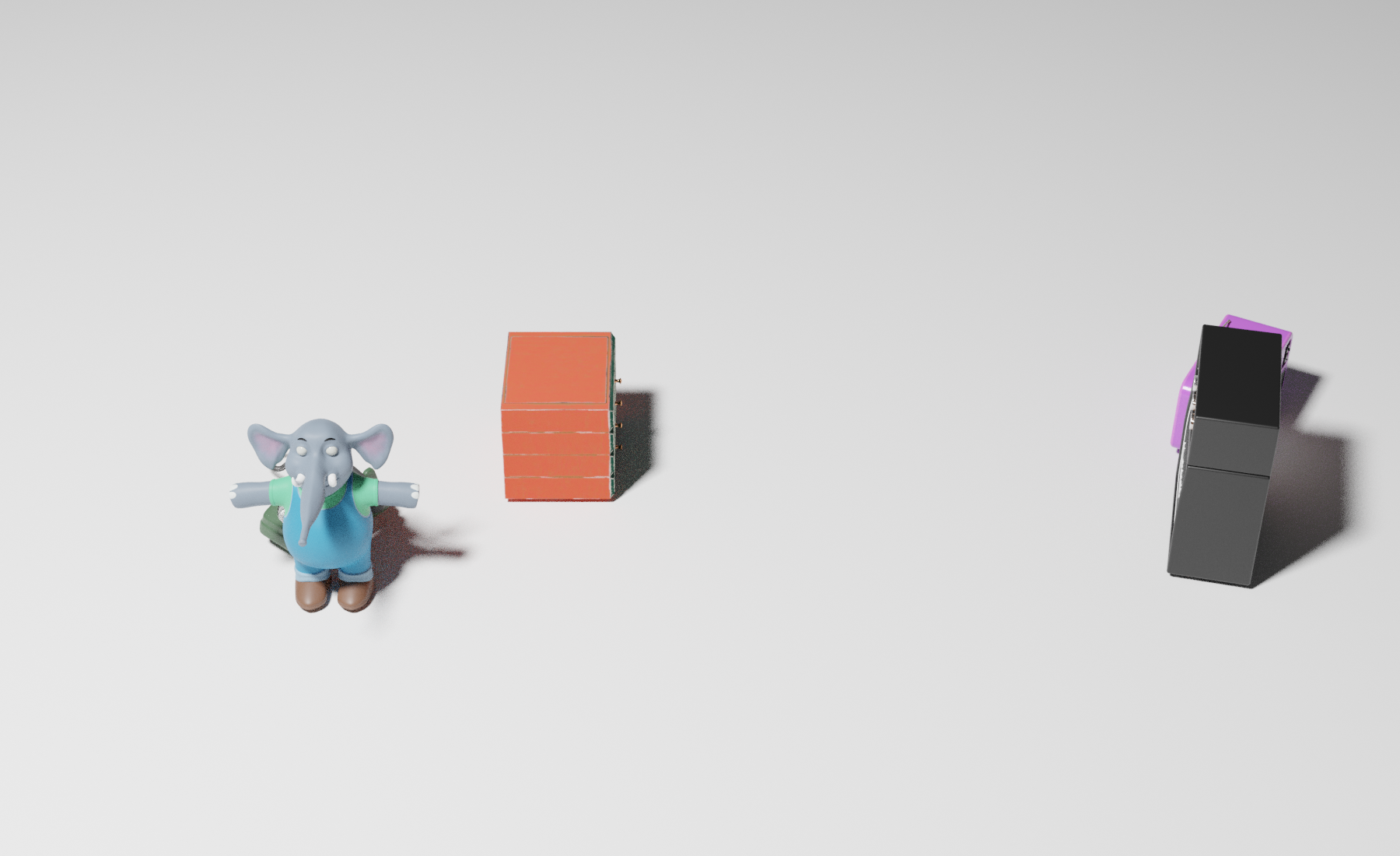}\hfill
    \includegraphics[width=0.497\linewidth]{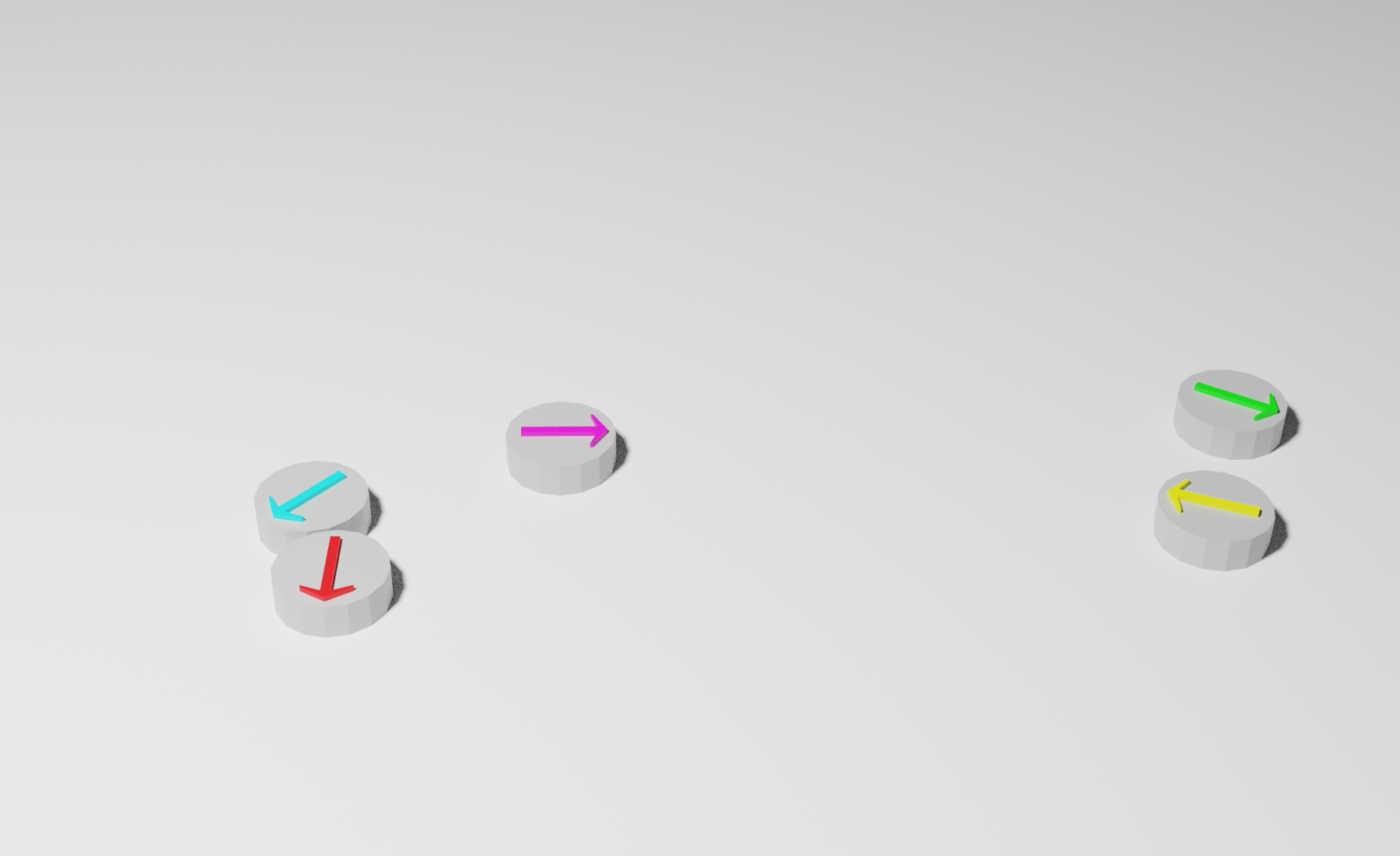}\hfill
    \includegraphics[width=0.497\linewidth]{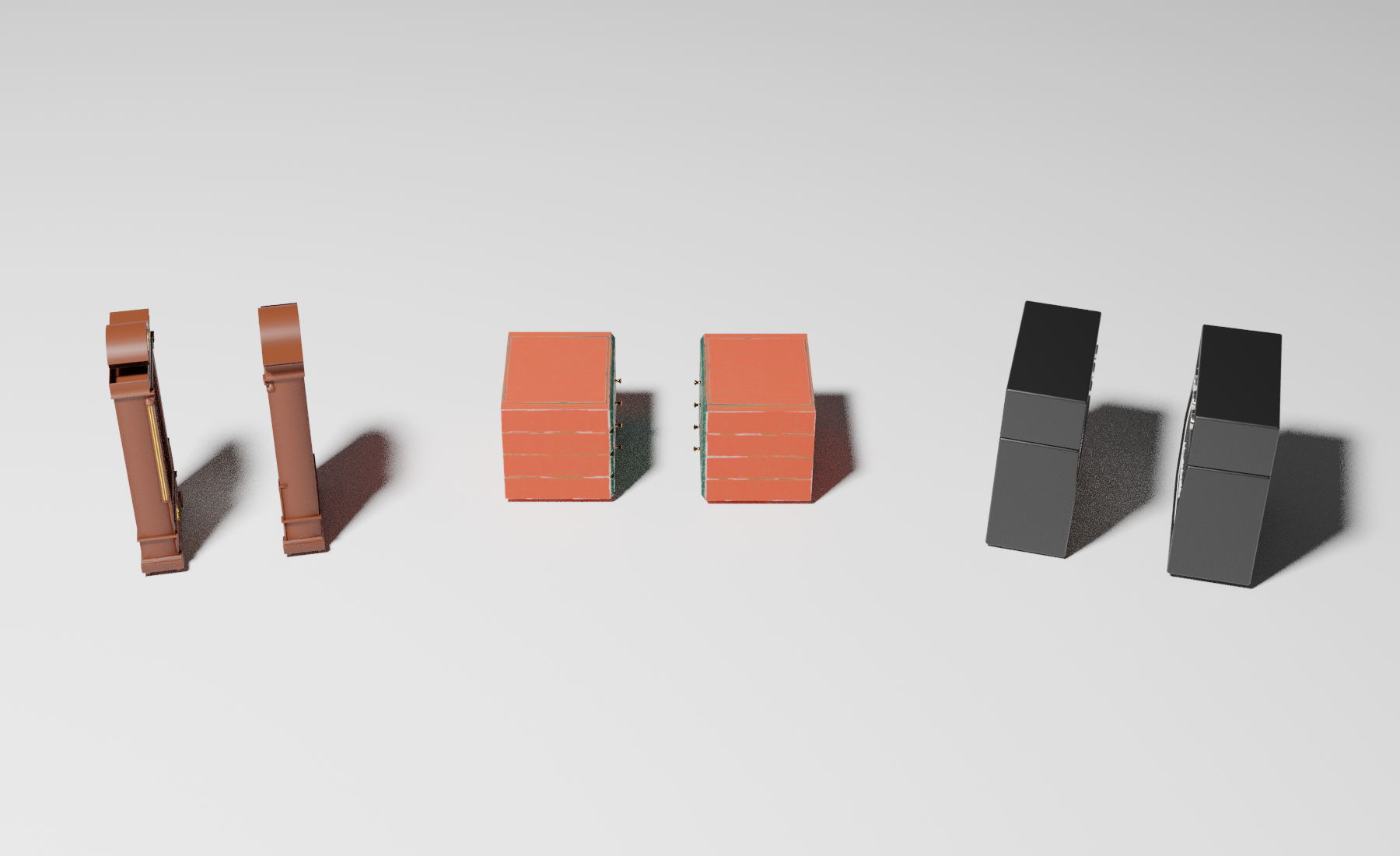}\hfill
    \includegraphics[width=0.497\linewidth]{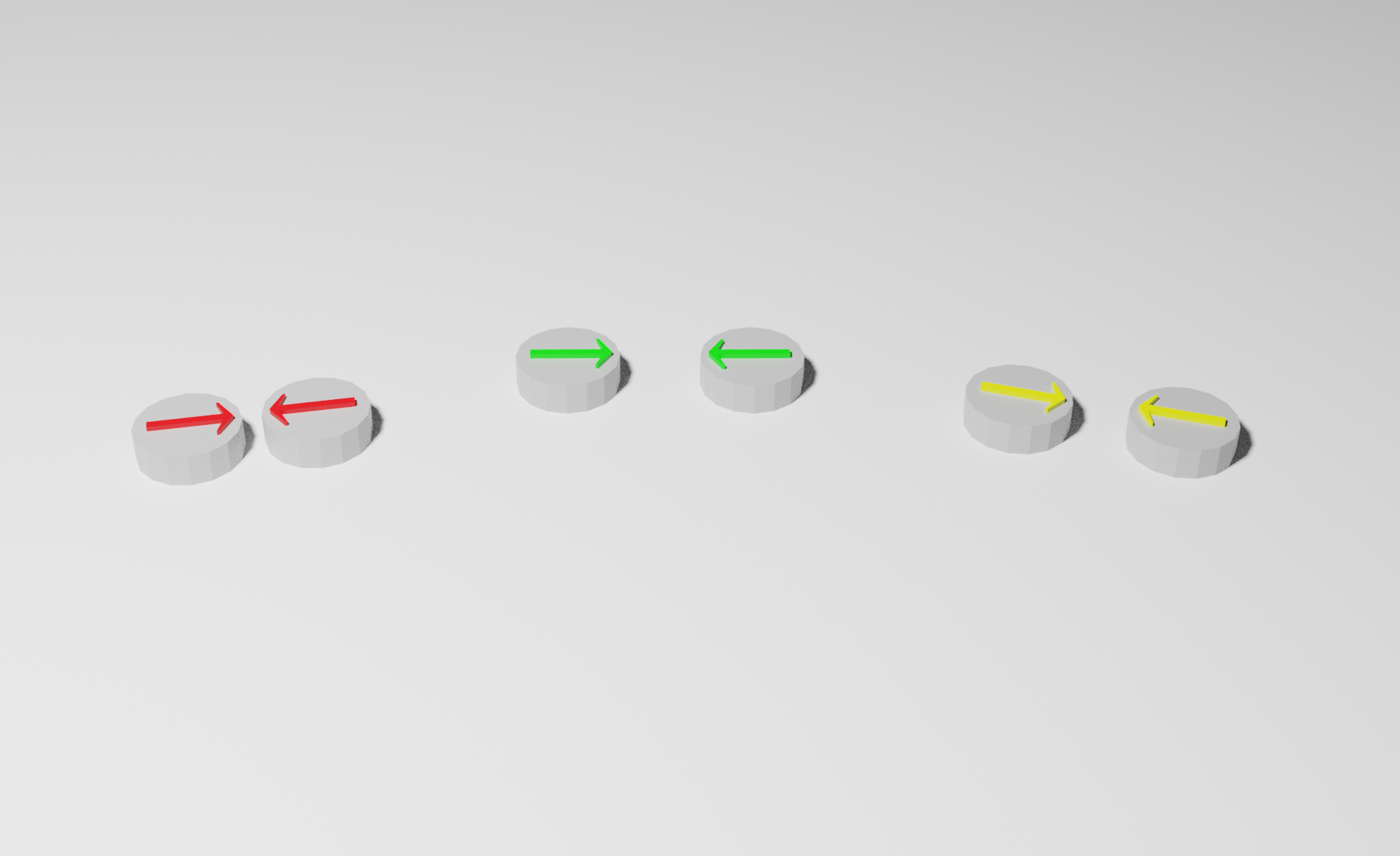}\hfill
       \caption{Common challenges in grounding relative directions arising with real objects, exemplified by objects from the original GRiD-3D dataset. \textbf{Top left:} Occlusion due to variability in heights and shapes of objects. \textbf{Bottom left:} Symmetry of objects impairs the detection of their front sides. \textbf{Top/bottom right:} Replica of the images on the left using abstract objects from the GRiD-A-3D dataset.}
   \label{fig:challenges}
\end{figure}


In this work, we aim to provide a variation of the original dataset that ensures the elimination of such potential distortions (see Fig.~\ref{fig:challenges} for examples), enabling a model to more quickly learn how to ground relative directions, which may be of particular value for few-shot, transfer, and curriculum learning scenarios. Accordingly, we extend the GRiD-3D benchmark suite towards another diagnostic VQA dataset with abstract objects. 



\section{GRiD-A-3D Abstract VQA Dataset}
With the introduction of the GRiD-3D dataset~\cite{lee_grid3d_2022}, we could show that neural VQA models are capable of grounding relative directions by implicitly deriving a curriculum of subtasks.
In order to further generalize the previous findings, we extend our GRiD-3D suite towards a diagnostic dataset based on abstract objects whose cardinal direction is indicated by colored arrows.

\paragraph{Overview and statistics}
With our new GRiD-A-3D dataset, we address the following six tasks:  \task{Existence Prediction}, \task{Orientation Prediction}, \task{Link Prediction}, \task{Relation Prediction}, \task{Counting}, and \task{Triple Classification}. All \mbox{8\,000} rendered images are split without overlap into \mbox{6\,400} for training, 800 for validation, and 800 for testing. The \mbox{432\,948} corresponding input questions follow largely the same 80:10:10 ratio, yielding \mbox{346\,984}, \mbox{43\,393}, and \mbox{42\,571} questions for each set, respectively. The GRiD-A-3D dataset has an order of magnitude comparable with the GRiD-3D dataset, both in terms of image and question counts.


\begin{wrapfigure}{r}{3cm}
\centering
\includegraphics[width=\linewidth]{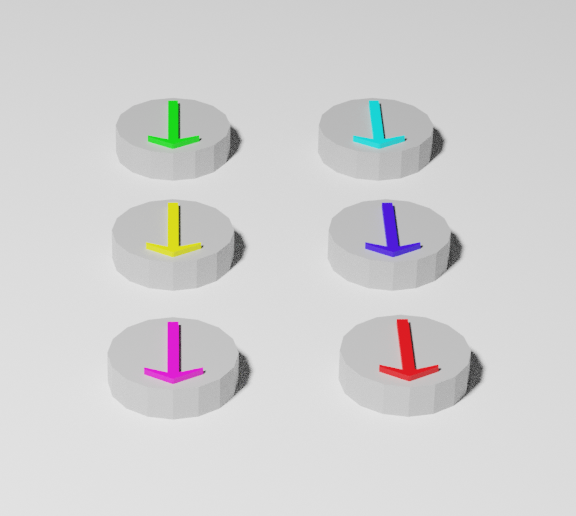}
\caption{The six abstract objects used in the GRiD-A-3D dataset.}
\label{fig:objects}
\end{wrapfigure}

\paragraph{Image generation}
For each image, we generate a scene by randomly placing three to five distinct objects onto a plane and render the corresponding image with 480x320 pixel resolution via Blender.\footnote{\url{https://www.blender.org/}} We choose a consistent lighting setup across all images, add shadows to each object, and restrict the image generation to a fixed camera angle, thus obtaining one image per scene.     

Our object set comprises gray-coloured polygonal prisms approximating a cylinder shape, each marked with an arrow in one of the six different colours: three primary colours (red, blue, and green) and three additive secondary colours (yellow, cyan, and magenta). 
The tip of each arrow depicts the object's front side, allowing for distinct relative directions between objects in the image. An overview of all six objects can be found in Fig.\ \ref{fig:objects}. Note that the overall object count in the original GRiD-3D dataset is 28, whereas GRiD-A-3D is restricted to six different objects.




\begin{figure}[h]
  \centering%
  \begin{subfigure}[t]{0.47\linewidth}
    \centering%
    \includegraphics[width=\linewidth]{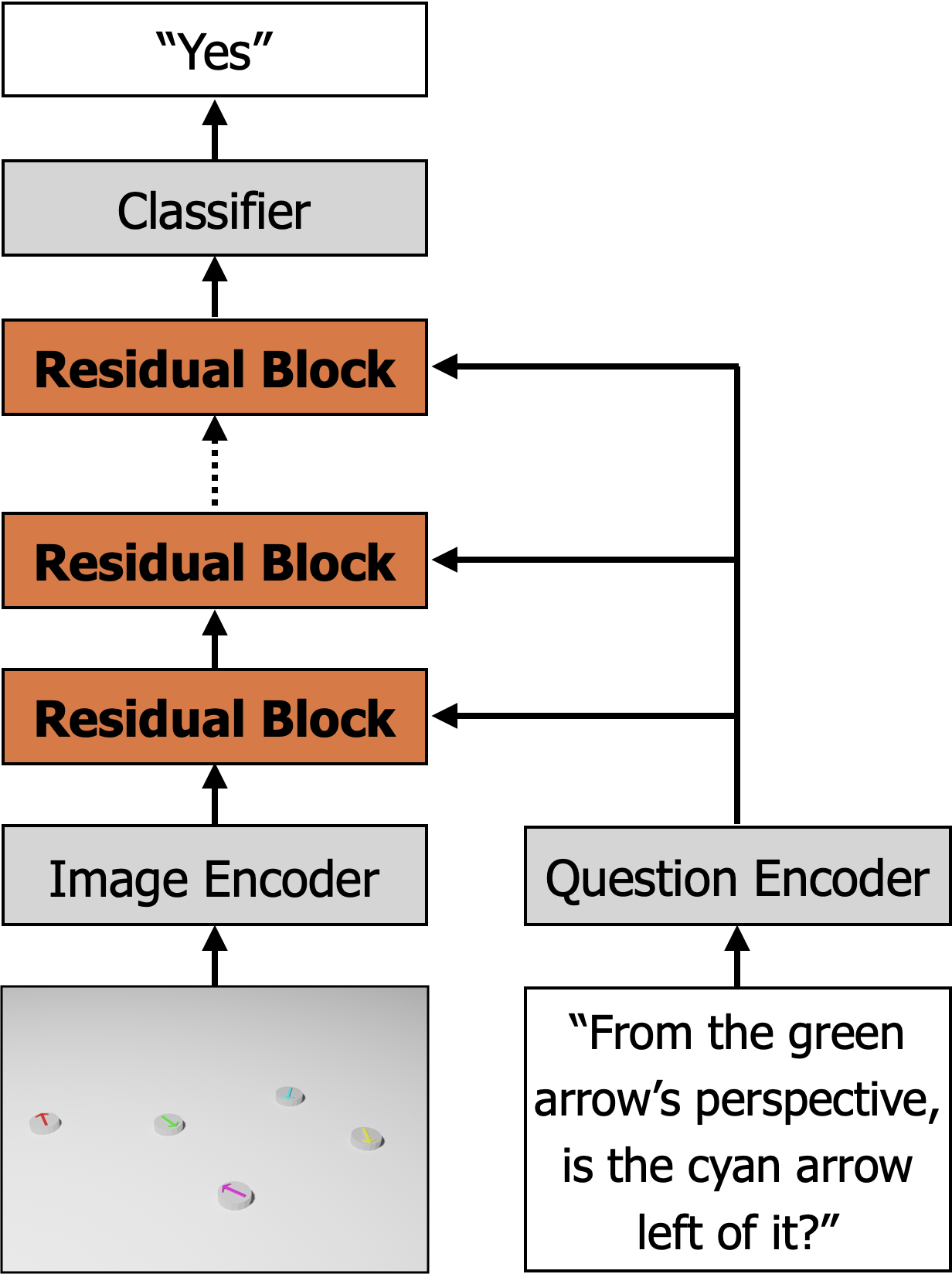}
    \caption{FiLM}
  \end{subfigure}
  \hfill
  \begin{subfigure}[t]{0.47\linewidth}
    \centering%
    \includegraphics[width=\linewidth]{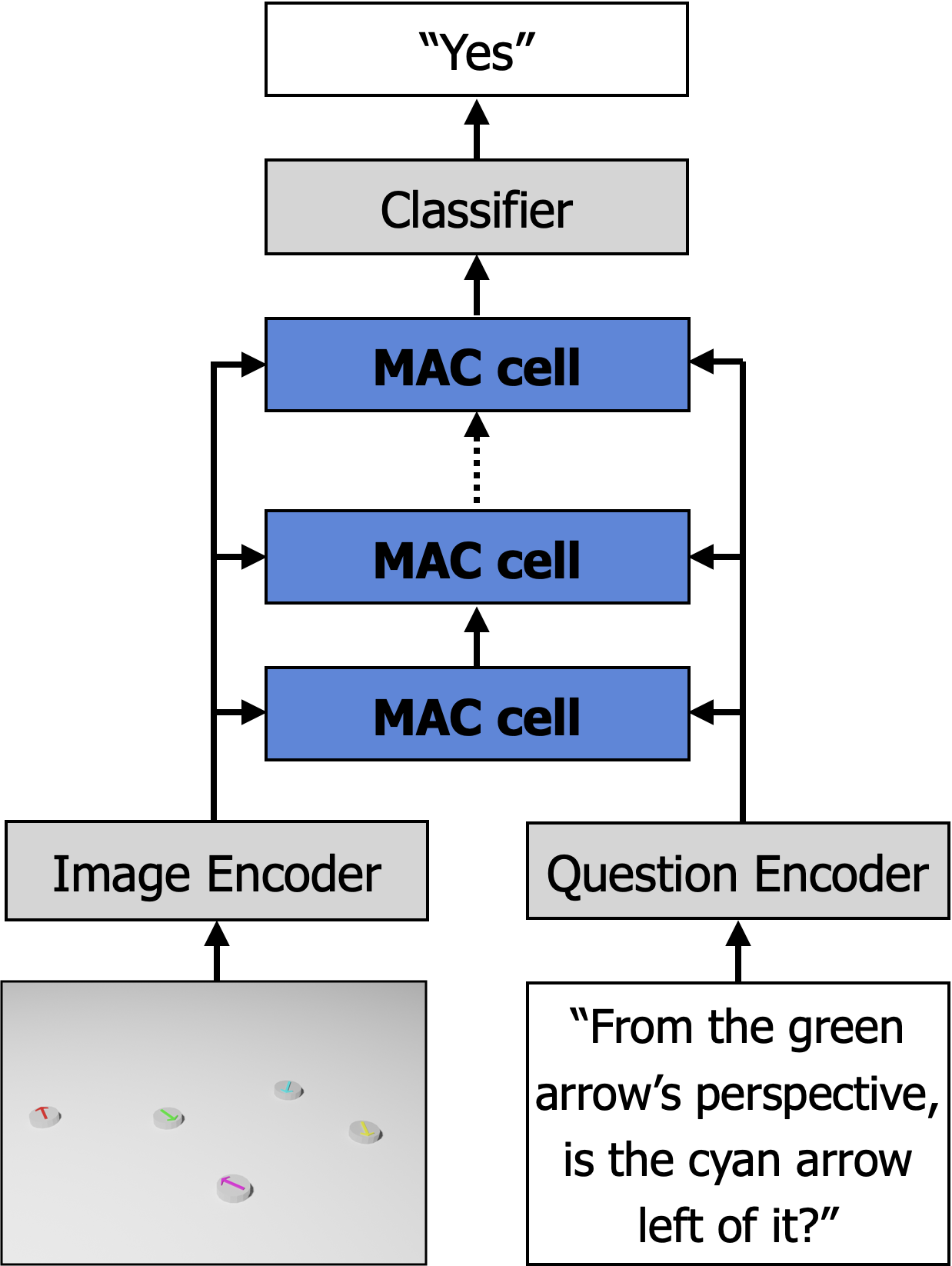}
    \caption{MAC}
  \end{subfigure}
  \caption{Neural end-to-end VQA models FiLM and MAC used for our experiments. The generic units (here colored in orange and blue, respectively) control how the question and image features are being processed. \label{fig:filmandmac}}
  
\end{figure}

\begin{figure*}[t!]
    \includegraphics[width=0.33\textwidth]{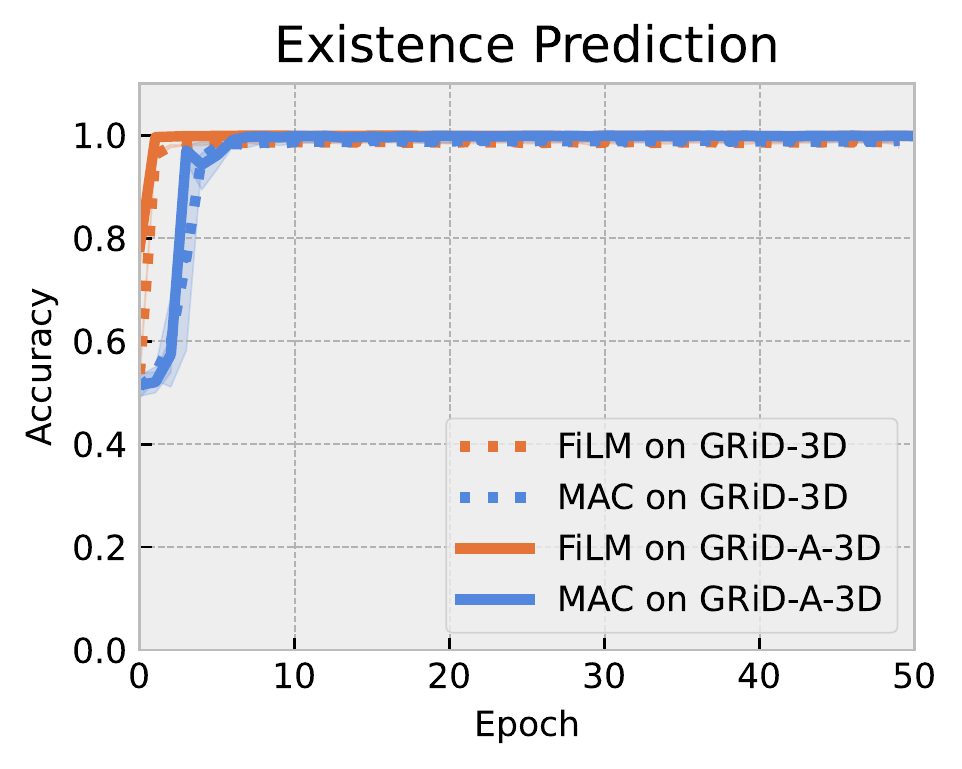}\hfill
    \includegraphics[width=0.33\textwidth]{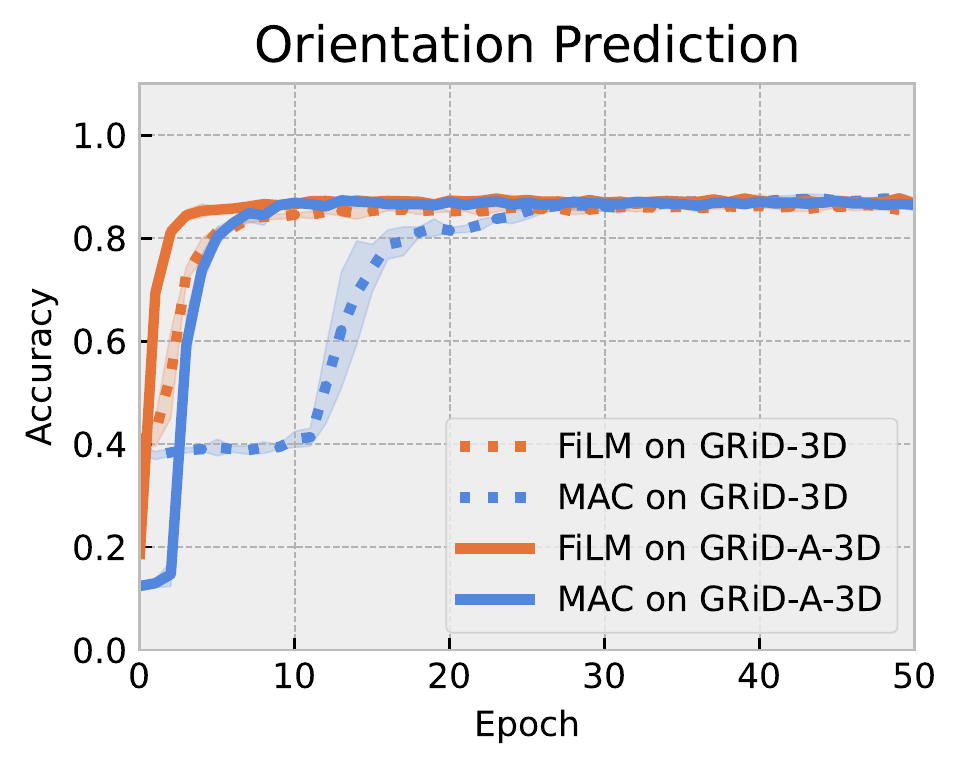}\hfill
    \includegraphics[width=0.33\textwidth]{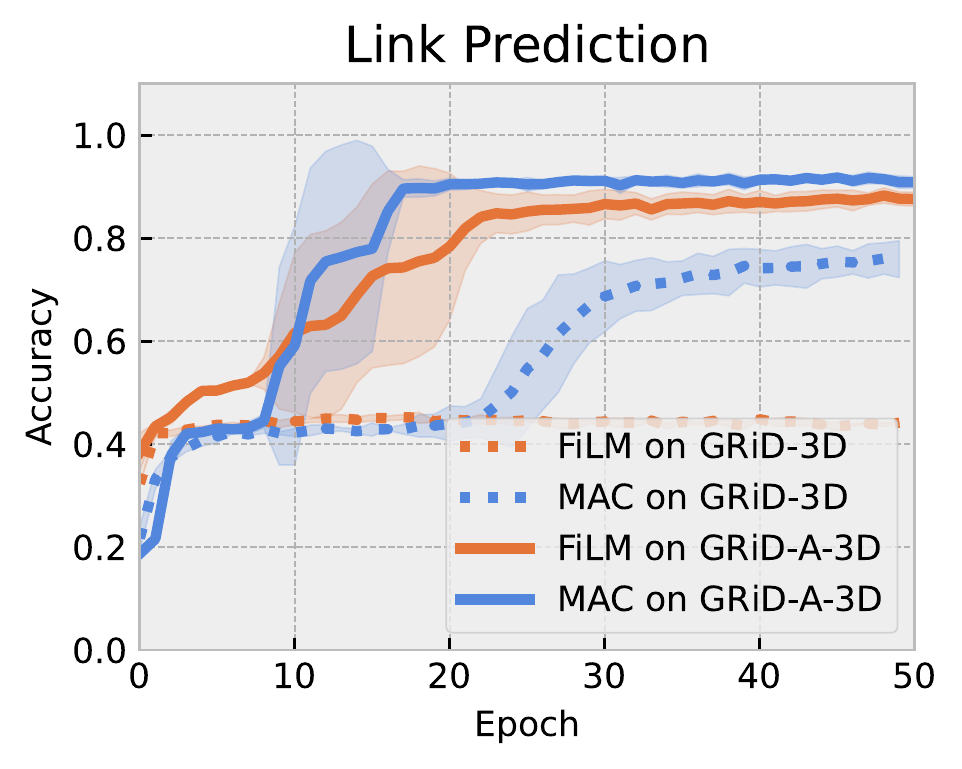}\\
    \includegraphics[width=0.33\textwidth]{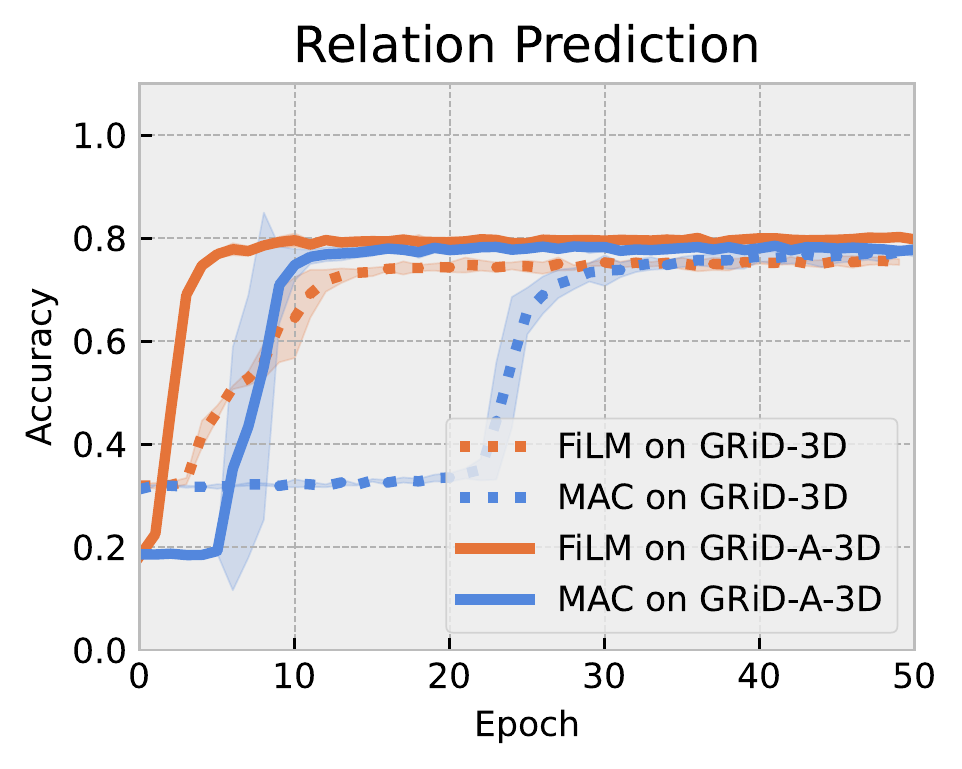}\hfill
    \includegraphics[width=0.33\textwidth]{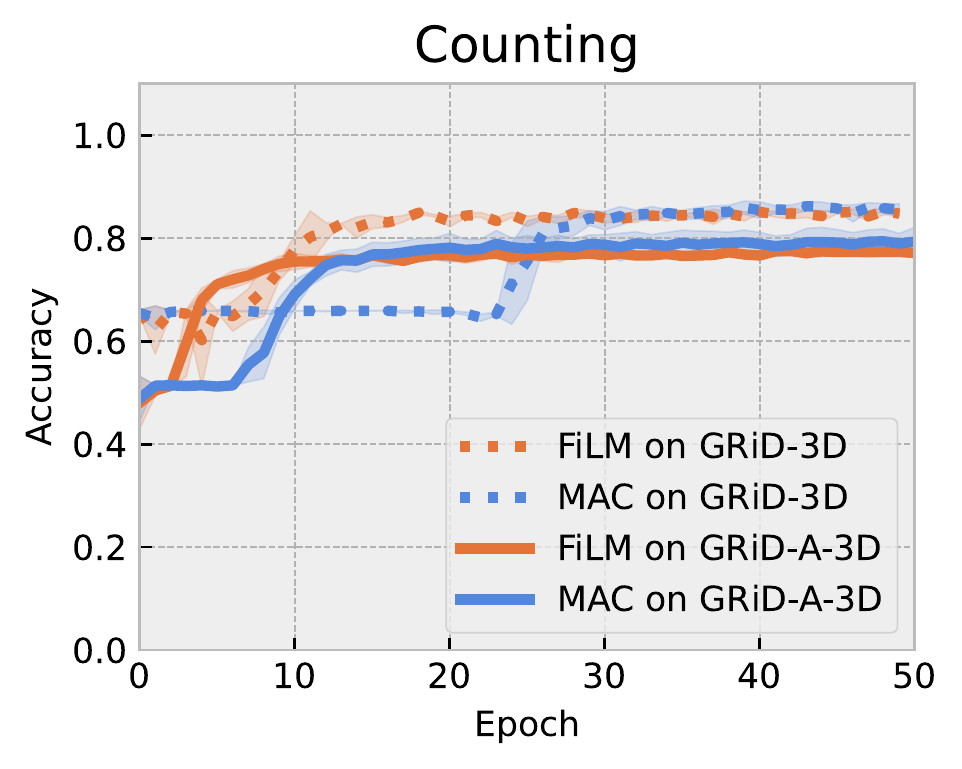}\hfill
    \includegraphics[width=0.33\textwidth]{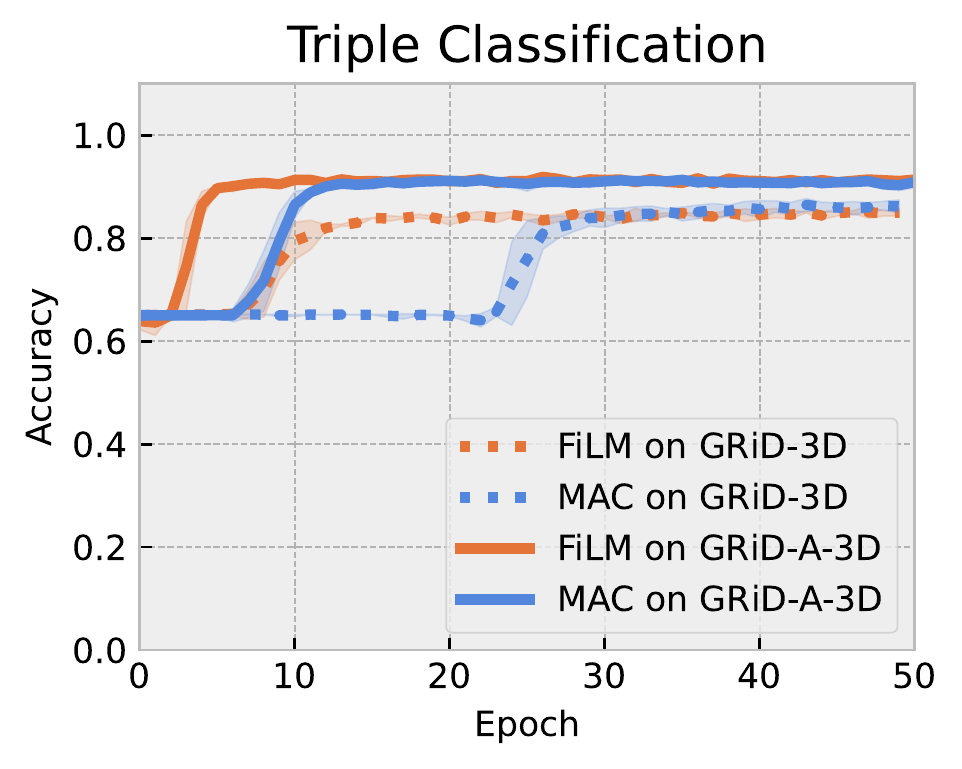}
    \caption{Multi-task learning results of FiLM (orange lines) and MAC (blue lines) on each of the six reasoning tasks of the GRiD-A-3D dataset (solid lines) vs. training the same models on the original GRiD-3D dataset (dotted lines).\protect
    \label{fig:learning-dynamics}}
\end{figure*}

\paragraph{Question generation}
In addition to rendering the images from our sampled scenes, we obtain scene graphs equipped with ground truth information such as absolute position, orientation, and relative directions of objects, that we use to generate questions related to the six tasks contained in GRiD-A-3D. Our question generation builds upon the framework provided with CLEVR~\cite{johnson_clevr_2017}, whose question templates, synonym, and metadata files we tailor to our dataset. Likewise, our question generation pipeline is expressed as a template-based functional program executed on each scene graph. 

We follow the depth-first search strategy to determine and instantiate question-answer pairs that comply with the scene information and can therefore be considered valid. We set additional constraints to make sure that answers are uniformly distributed for each task. To ensure a wide variety of natural language questions, we sample from a rich set of differently phrased question templates for each reasoning task and randomly omit utterances or replace words with suitable synonyms. 
 
\section{Evaluations}
\label{sec:evaluations}

For our experiments, we train MAC~\cite{hudson_compositional_2018} and FiLM~\cite{perez_film_2018}, two state-of-the-art neural end-to-end VQA architectures, on our new GRiD-A-3D dataset (cf.~Fig.~\ref{fig:filmandmac}). Both architectures take raw RGB images and plain text question-answer pairs as input for training.
Image features are extracted by a pretrained ResNet101~\cite{he_resnet_2016} for both models, while questions are encoded by a GRU~\cite{chung_gru_2014} (FiLM) or a bidirectional LSTM~\cite{hochreiter_lstm_1997} (MAC), respectively.  
Subsequently, image and question features are fed to special neural units called \emph{residual blocks} (FiLM) or \emph{MAC cells} (MAC). A chain of such units provides the core of the reasoning process.     


We use existing PyTorch\footnote{\url{https://pytorch.org/}} implementations of FiLM and MAC with their default hyperparameters for the published CLEVR~\cite{johnson_clevr_2017} dataset evaluations, except for the number of MAC cells that we reduce to four to prevent overfitting. All experiments are run for 100 epochs and repeated three times with different seeds to reduce the impact of the random initialization of the models on the results. Fig.\ \ref{fig:learning-dynamics} shows the \emph{mean} and the \emph{standard deviation} of the evaluations.



We interpret our results in the following way: \task{Existence} and \task{Orientation Prediction} are learned earlier than other tasks. We explain this observation with the fact that these tasks only require a model to focus on one single object. For the most straightforward task of \task{Existence Prediction}, we observe similar behavior for the two datasets: Both converge to an accuracy of almost 100\% at nearly the same time. For the \task{Orientation Prediction} task, we observe convergence to an accuracy of over 80\% for both datasets. Noticeably, the learning happens faster for the abstract GRiD-A-3D dataset. The shorter learning time can be attributed to the more unequivocal identification of front and back sides of the abstract objects due to the lack of symmetry related noise as shown in Fig.~\ref{fig:challenges}. The fact that the accuracy on \task{Orientation Prediction} is capped at about 85\% can be explained by objects placed close to the border between two cardinal directions, as such cases are difficult for the models to learn and classify.

A similar learning behavior can be observed for the more complex tasks of \task{Relation Prediction}, \task{Triple Classification} and \task{Link Prediction}, where both models converge faster when trained on GRiD-A-3D and also reach slightly higher accuracy. Similarly to the results on the \task{Orientation Prediction} task, the main reason for these observations may lie in the facilitated learning conditions due to the lack of front-back symmetries or strong occlusions with the abstract objects.
This effect is most pronounced for \task{Link Prediction}, i.e., predicting which target object is in a given relation to some reference object. We attribute this observation to the smaller set of objects in the GRiD-A-3D dataset.

Finally, we observe a mixed result for the \task{Counting} task: While learning of both VQA models converges faster for the GRiD-A-3D dataset, higher accuracy is reached for the GRiD-3D dataset. We hypothesize that this higher accuracy stems from the more diverse-looking objects in the GRiD-3D dataset, facilitating the models to distinguish and thus count multiple objects in close proximity.

In summary, our results suggest the following two facts: First, the abstract GRiD-A-3D dataset leads to  faster learning and can thus enable more computationally efficient experimentation while achieving comparable results to the original GRiD-3D dataset. Second, the results support our assumption of a chronology of subtasks, as \task{Existence Prediction} and \task{Orientation Prediction} are learned before the models can reason about relative directions.

\section{Conclusions}
This work is an extension to previous work on grounding relative directions with end-to-end neural VQA architectures.
We provide a comprehensive, simplified GRiD-A-3D dataset with abstract objects that shows similar behavior to the original GRiD-3D dataset when learned by the two established VQA models FiLM and MAC. With our experiments, we show that the learning of tasks that focus on a single object like object recognition and orientation prediction happens prior to learning to ground relative directions and object counting.

The abstract nature of the dataset eliminates approximate front-back object symmetries that can have a negative impact on object orientation prediction and all reasoning tasks about directional relations that build upon it. Furthermore, the simplification of the object set allows for conducting experiments with a more comprehensive dataset. In future work, this will allow us to conduct fast pilot studies on curriculum and transfer learning based on the intuitive dependency of the different spatial reasoning tasks on one another and the observed implicit curriculum.



\section*{Acknowledgments}
The authors gratefully acknowledge support from the German Research Foundation DFG for the projects CML TRR169, LeCAREbot and IDEAS.

\bibliographystyle{named}
\bibliography{jl,mk,ka}%
\end{document}